# Session-based Cyberbullying Detection in Social Media: A Survey

Peiling Yi[a,*], Arkaitz Zubiaga[a]

[a]School of Electronic Engineering and Computer Science, Queen Mary University of London, Mile End Road, London, E1 4NS, United Kingdom

ARTICLE INFO

*Keywords*:
Cyberbullying detection
Cyberbullying
Session-based cyberbullying detection
Social media
Machine learning
Natural language processing

ABSTRACT

Cyberbullying is a pervasive problem in online social media, where a bully abuses a victim through a social media session. By investigating cyberbullying perpetrated through social media sessions, recent research has looked into mining patterns and features for modelling and understanding the two defining characteristics of cyberbullying: repetitive behaviour and power imbalance. In this survey paper, we define the Session-based Cyberbullying Detection framework that encapsulates the different steps and challenges of the problem. Based on this framework, we provide a comprehensive overview of session-based cyberbullying detection in social media, delving into existing efforts from a data and methodological perspective. Our review leads us to proposing evidence-based criteria for a set of best practices to create session based cyberbullying datasets. In addition, we perform benchmark experiments comparing the performance of state-of-the-art session-based cyberbullying detection models as well as large pre-trained language models across two different datasets. Through our review, we also put forth a set of open challenges as future research directions.

## 1. Introduction

Cyberbullying is a form of bullying that is perpetrated through online devices [1]. "Bullying" is defined as the repeated and deliberate aggressive behaviour by a group or individual towards a person who is in a more vulnerable position to defend themself [2]. While the precise definition of cyberbullying varies slightly across studies, there are two characteristics that are consistently referred to: repeated aggression and power imbalance, both of which are key aspects to identify cases of cyberbullying behaviour [3, 4, 5, 6].

Figure 1 illustrates an example of an act of cyberbullying in an online chat [7, 8]. A bully recurrently sends mocking messages exposing personal or sensitive information of an indefensible victim. Messages by both the victim and the bully may contain offensive words, where however the victim will generally be using such words to try to defend themself from the bully.

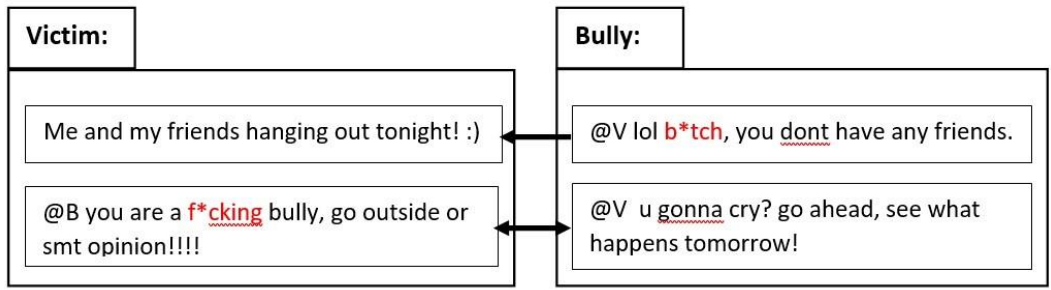

**Figure 1:** Example of a case of cyberbullying.

Cyberbullying detection is the task of automatically identifying cyberbullying events from online data, with the aim of stopping the abuse and preventing further harm [9, 10]. Developing an ability to detect cyberbullying events is however challenging, as it needs to capture the recurrent nature of the abusive behaviour. An ability to detect offensive

---

*Corresponding author

p.yi@qmul.ac.uk (P. Yi); a.zubiaga@qmul.ac.uk (A. Zubiaga)

http://www.zubiaga.org/ (A. Zubiaga) ORCID(s): 0000-0003-4583-3623 (A. Zubiaga)

or toxic sentences, as in for example hate speech detection and abusive language detection [11], does not suffice; ideally, it needs to consider the full history of the conversation (i.e. a social media session) to identify the recurrent nature inherent to cyberbullying events [12], by constructing a representation of the interaction between the bully and the victim.

In this survey paper, we delve into the current development of research into cyberbullying detection in social media, with a particular focus on methods incorporating social media sessions into their pipeline. A number of recent surveys have covered cyberbullying detection, which have however had a different focus from the one here on social media sessions. Existing surveys have predominantly covered cyberbullying detection in general without a focus on social media sessions [13, 14, 15, 16, 17], as well as others have focused on more specific aspects including a critical review on the definitions and operationalisation of cyberbullying [18], an overview of the implications of cyberbullying [13] and providing a taxonomy of the different types of cyber-attacks [19].

There are important aspects that a social media session can provide that cannot be inferred from isolated posts. Performing cyberbullying detection by modelling social media sessions provides a holistic view into the power imbalance between the bully and the victim, which is not as clear from isolated posts. In addition, the repetitive nature of cyberbullying can only be captured by the sequence of comments in a conversation session.

We survey 10 publicly accessible cyberbullying datasets and 55 cyberbullying detection models by examining how they adhere to the conditions set out above, which we formalise in a cyberbullying detection framework. We refer to this framework as the Social Media Session-Based Cyberbullying Detection (SSCD) that unifies the definition, data collection and detection of cyberbullying.

Our survey paper makes a number of contributions, of which we highlight:

- We define the four steps of SSCD: (i) Social media platforms selection, (ii) Session-based data collection, (iii) Cyberbullying annotation and (iv) Session-based cyberbullying detection.

- We give an overview of the existing datasets and methods in accordance with the SSCD framework.

- We define a set of evidence-based criteria recommended for the selection and creation of a SSCD dataset.

- We perform experiments investigating the use of two state-of-the-art session-based cyberbullying detection models and eight different pre-trained language models to tackle SSCD tasks.

- Informed by our literature review on existing datasets and methods, we provide a set of suggestions for consideration in future work for dataset creation, model development as well as reporting in scientific publications.

## 1.1. Paper structure

The aim of the survey is to provide a comprehensive understanding of cyberbullying detection at the social media session level from a data and methodological perspective. In the next section, we introduce and describe the SSCD framework that defines the modelling of the cyberbullying detection task considering social media sessions. Then, in section 3, we describe the methodology we follow to conduct this survey as well as how the relevant research papers were selected. Section 4 discusses existing datasets, for which we analyse 10 cyberbullying detection datasets, discussing their data collection and annotation strategies. In Section 5 we then discuss existing methods and research directions in cyberbullying detection as well existing efforts on modelling the task in line with the SSCD framework. We continue by presenting experiments with state-of-the-art language models in Section 6. After studying datasets, models and experiments, we provide a set of best practices for conducting SSCD research in Section 7. After that, we discuss open challenges within session-based cyberbullying detection in Section 8 before concluding the paper in Section 9.

## 2. SSCD Framework

Social media sessions are ubiquitous ecosystems of cyberbullying. Hence, in line with recent efforts in cyberbullying detection [20, 12], we argue for the need to consider the two key characteristics of cyberbullying (i.e. repetitive behaviour and power imbalance) in modelling the cyberbullying detection task. We operationalise this through what we name the Social Media Session-Based Cyberbullying Detection (SSCD) framework. Figure 2 illustrates the structure of the SSCD framework, which consists of four main steps:

1. Social media platform selection: the starting point consisting in choosing the social media platforms to be considered in the data collection.
2. Session-based data collection: it differs from collection of individual posts in that the unit being collected is an entire social media session, and so is the unit that is annotated for cyberbullying detection. One may also distinguish two types of social media sessions: (i) conversation sessions, which are text-based sessions involving at least two users, as in Figure 1, and (ii) media sessions, which include other types of media beyond just text, as in Figure 3.
3. Cyberbullying annotation: where it is crucial to provide detailed and clear criteria for defining what constitutes a case of cyberbullying.
4. Session-based cyberbullying detection: where the model for cyberbullying detection is built and evaluated.

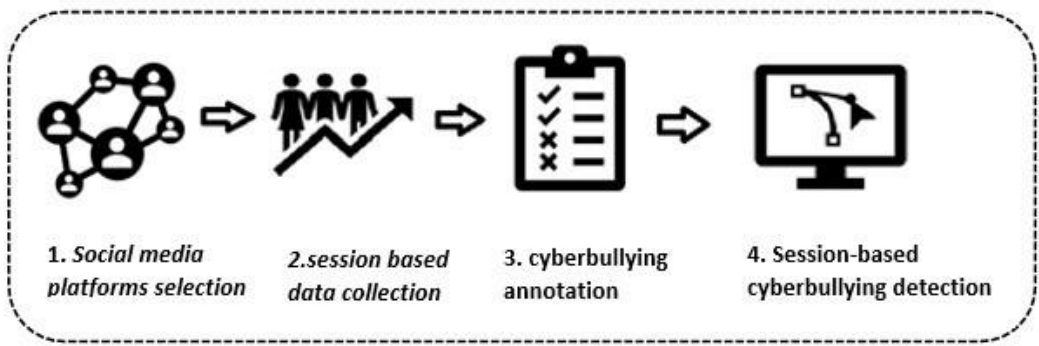

**Figure 2:** A general framework for social media session-based cyberbullying detection (SSCD).

Figure 3 illustrates an example of SSCD, based on the collection strategy followed by [21], who constructed a media session-based cyberbullying dataset on Vine, a video-sharing platform. Cyberbullying can happen on Vine in a number of ways, such as posting offensive comments, re-editing or transcribing someone's video for mockery. The collection and study of video-based social networking sessions was first done by [21]. They defined a social media session in Vine as the posting of a video with its associated likes and comments, restricting collection in this case to a minimum of 15 comments in order for the annotator to have enough context to assess the frequency/repetition of profanity and imbalance power that fits the definition of cyberbullying. Annotators were trained prior to their participation, and were given clear instructions explaining the distinctions between cyberaggression and cyberbullying along with a sample of media sessions.

Throughout the paper, we will refer back to the SSCD framework in Figure 2, linking to the relevant parts.

## 3. Survey Methodology

### 3.1. Selection of studies

We followed a consistent methodology to retrieve relevant papers to be covered in our survey, with the aim of ensuring good coverage of papers while also avoiding biased selection based on subjective criteria. We used four different keywords (i.e. 'cyberbullying detection', 'bullying detection', 'cyberbullying datasets', 'cyberbullying software') to retrieve publications from a wide set of scientific search engines, including Google Scholar, ACM, IEEE, arXiv. We then reviewed the included studies according to the 2020 PRISMA Update Study Flowchart [22]. The process is shown in Figure 4.

After retrieving all these papers and removing duplicates, 217 publications were selected for more careful analysis and validation. A final set of 55 was selected after removing publications that didn't fall into any of the following exclusion criteria:

• Exclusion criteria 1: Cyberbullying detection is not applied on social media platforms, hence it's outside of our scope.

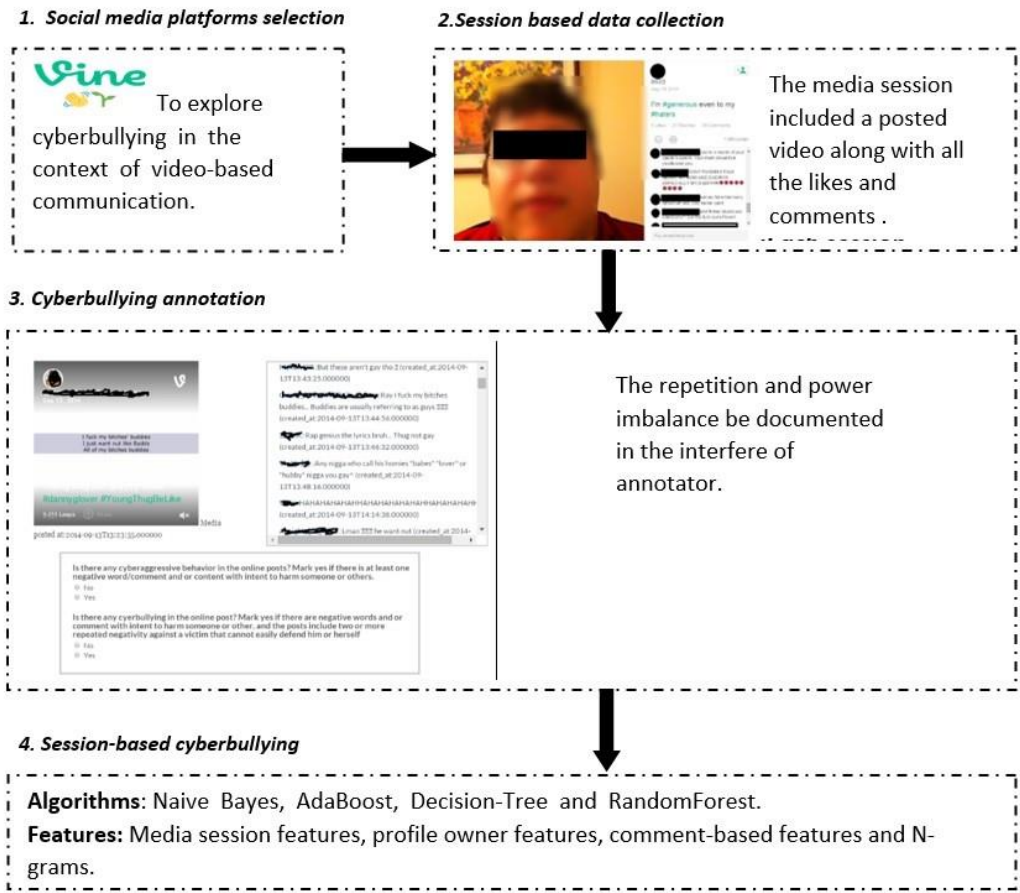

**Figure 3:** An instance of SSCD framework [21].

• Exclusion criteria 2: While the study is about or mentions cyberbullying detection, there is no implementation, study or evaluation of a detection model, e.g. papers discussing how cyberbullying detection can inform policy.

• Exclusion criteria 3: The study is a review, survey, or a study of theoretical concepts about cyberbullying or cyberbullying detection, without any empirical implementation or analysis.

Some of the publications matching exclusion criteria 2 or 3 are either discussed throughout this paper or cited for backing up some of our statements, however they don't conform the set of studies used for our core discussion of datasets and models for cyberbullying detection.

## 3.2. Approach for study analysis

The main aim of this survey paper is to provide a comprehensive understanding of SSCD from the perspectives of data and methodologies, and combining both to perform a critical analysis of SSCD. To do so, in this study we adopt the "data statements" framework [23]. Data statements suggest how datasets should be created in Natural Language Processing (NLP) research, with the aim of increasing transparency and helping alleviate issues related to exclusion and bias [24]. We also follow guidelines provided by Kitchenham [25] for writing systematic reviews on the subject of software engineering, which we adapt to the field of NLP and cyberbullying detection.

## 4. Datasets

Among the publications considered within this survey, we selected all those with publicly available or otherwise accessible (e.g. by contacting authors) datasets, the ones that enable further research as well as allow to explore the

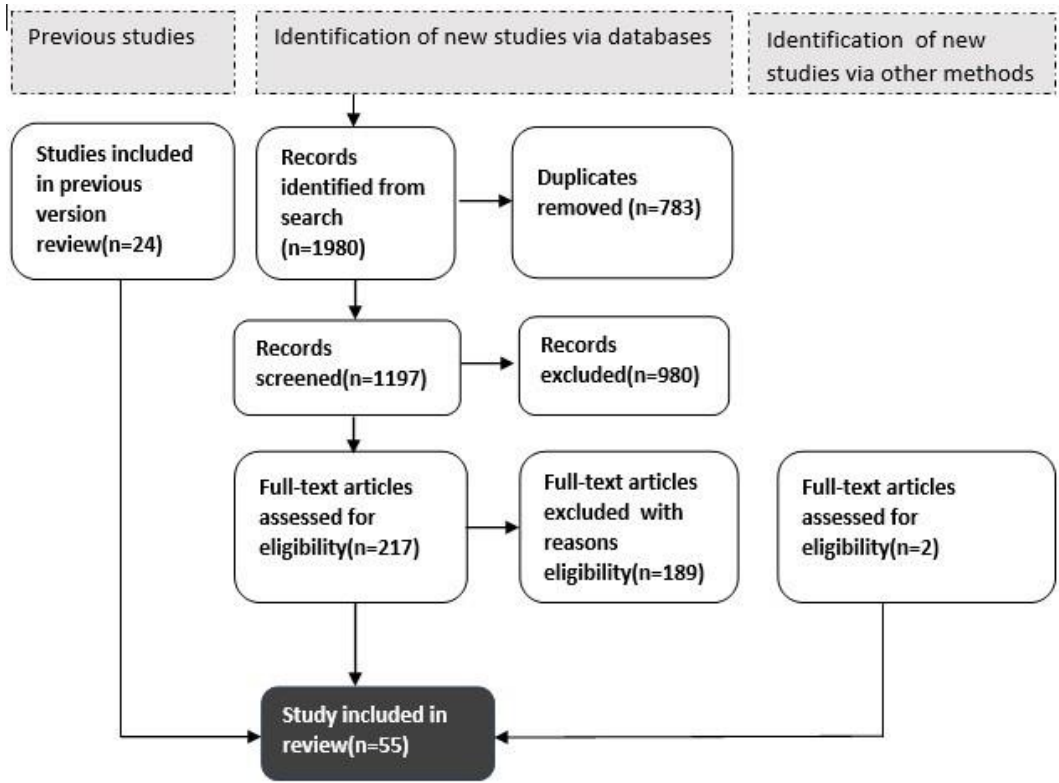

**Figure 4:** PRISMA Model for Cyberbullying Detection Research.

| Dataset statistics | | | | | | | SSCD Datasets Criteria | |
|---|---|---|---|---|---|---|---|---|
| Platforms | Size | Ration | Year | Annotation | Collection | Source | Session_based | Rigorous Definition |
| FormSpring [26] | 12,773 | 0.08 | 2011 | Crowd-sourcing | Crawled | chatcoder.com | Yes | No |
| Myspace [27] | 4,813 | 0.21 | 2011 | Research assistant | Crawled | chatcoder.com | Yes | Yes |
| YouTube[28] | 3,468 | 0.14 | 2014 | Research assistant | Crawled | Figshare.com | No | No |
| Instagram [29] | 2,218 | 0.29 | 2015 | Crowd-sourcing | Crawled | cucybersafety.org | Yes | Yes |
| Vine [30] | 970 | 0.31 | 2015 | Crowd-sourcing | Crawled | cucybersafety.org | Yes | Yes |
| Twitter [31] | 534,950 | 0.29 | 2015 | N/A | Twitter's API | chatcoder.com | No | No |
| Wikipedia [32] | 115,864 | 0.11 | 2017 | Crowd-sourcing | Crawled | github.com/sweta20 | Yes | No |
| Twitter [33] | 16,090 | 0.32 | 2017 | Crowd-sourcing | witter's API | github.com/sweta20 | No | Yes |
| ASKfm [8] | 90,296 | 0.15 | 2018 | Crowd-sourcing | Crawled | cucybersafety.org | Yes | No |
| Twitter [34] | 47,000 | 0.16 | 2020 | N/A | Twitter's API | kaggle.com | No | No |

**Table 1**

Available cyberbullying datasets. Datasets collected based on social media sessions and following a rigorous definition of cyberbullying are highlighted with a grey background.

datasets. We gathered a total of 10 datasets that we discuss here. We present key information and statistics for each of these datasets in Table 1. In addition to general information such as dataset size, source and annotation methodology, we also include two additional tables based on SSCD criteria: (i) whether the dataset was collected with social media sessions as units, and (ii) whether the data collection followed a rigorous definition of cyberbullying, based on the description of the paper in question. Datasets adhering to both these criteria are highlighted with a grey background in the table. Looking at the statistics and characteristics of these datasets, we make a few observations next:

*Datasets are diverse.* If we look at the type of data and source platform used to collect the datasets, we see that they are rich in diversity. Across the 10 datasets, as many as 8 different social media platforms were used as sources, where the only platform with more than one dataset is Twitter. The main benefit of this is that it enables further investigation into the problem across different platforms, enabling in turn developing more generalisable models that can detect acts of cyberbullying in different environments. However, there has been little effort to develop more of these datasets in recent years (e.g. 8 datasets were created between 2011-2017, whereas only 2 from 2018-2022).

*Datasets are generally imbalanced.* Eight of the datasets have a class imbalance where fewer than 30% of the samples belong to the cyberbullying class. This imbalance is a challenge as it has been widely shown to affect the predictive power of machine learning classifiers [35] as shown on experimental reports from previous studies [36, 5, 37].

*Varying dataset sizes.* Dataset sizes vary significantly, from the Vine dataset containing 970 samples, to one of the Twitter datasets containing over 534K samples. While datasets generally contain over 10K samples, there is a clear difference in size for the session-based datasets, containing 970, 2.2K and 4.8K samples, which are understandably smaller given the increased cost of labelling entire sessions.

*Limited availability of SSCD datasets.* According to our analysis following the SSCD framework, only 6 of the datasets are collected based on sessions, and 4 datasets are labelled following a rigorous definition of cyberbullying. Overall, only 3 of the datasets satisfy both criteria.

*Mismatch between reported and published datasets.* The dataset size as stated in the original paper statement doesn't always match the size of the available dataset. What we report here is the size of the available dataset.

*Predominantly crowdsourced annotation of datasets.* The majority of the datasets (at least 6 out of 10) used crowdsourcing to annotate the datasets, with only two datasets relying on research assistants with expertise on the subject. Where expertise is important for a difficult annotation task like this, this calls for more datasets using trained annotators.

## 4.1. Selection of social media platforms

The starting point for constructing a SSCD dataset is the selection of a social media platform. This selection can be motivated by the objectives of the research, which can in turn inform how and what data to collect from the platform of choice, as well as the annotation instructions to provide to annotators.

In existing research, we observe that there have been predominantly two main reasons that motivated the choice of a social media platform:

- Platforms that are prone to cyberbullying events: Given the difficulty of retrieving cyberbullying events (i.e. a type of event that can be considered overall rare if we look at all the content in a platform), researchers often turn to platforms that are known to more frequently experience cyberbullying events. The choice of a platform on this basis can be motivated, for example, by the proportion of adolescents and college students known to be users of the platform [26, 29, 38]. Another feature is the anonymity allowed by certain platforms, which can also indicate higher presence of cyberbullying events in the platform [26]. This is the case of Formspring.me, a Q&A platform where users invite others to ask and answer questions anonymously.

- Platforms with no existing / public datasets: Another motivation to choose a particular social media platform has likely been the lack of existing datasets for that particular platform, which has led to a relatively diverse set of datasets from different platforms, as discussed in the previous section. This was for example the motivation of [32], who proposed to study personal attacks for the first time on Wikipedia, whereas [33] proposed to look at cyberbullying in short texts, hence focusing on Twitter. Others aimed to focus on cyberbullying events involving media content [29, 30, 28], which required exploration of new platforms.

## 4.2. Session-based data collection

Our exploration of datasets shown in Table 1 shows that 6 of the datasets followed a session-based data collection strategy. This makes it possible to more comprehensively capture the inherent feature of cyberbullying, i.e. repetitive behaviour. It also provides richer data representations that allow models to capture higher-level features. To delve deeper into the use of sessions in cyberbullying datasets, we next discuss datasets created according to two types of sessions: conversation sessions and media sessions.

*Conversation sessions.* In a typical conversation session, each item presents one question, followed by answers with their associated timestamps [26, 38, 27]. To fully understand the interactive nature of cyberbullying, the authors of [27] created a MySpace dataset with a moving window to capture each session.

*Media sessions.* Examples of media-based social networks include Instagram and Vine, where cyberbullying events are perpetrated through media-based communication. [29] collected a large sample of Instagram data, including 3,165,000 media sessions (images and their associated comments) from 25,000 user profiles, of which they labelled a small sample. [30] collected a dataset from the Vine platform, where each post is associated with a video, as well as a collection of likes and comments. While they collected over 650K media sessions, a small sample of it was labelled.

*No session-based.* The rest of the datasets are not collected based on social media sessions. They are generally made of isolated posts. Authors of [33] collected multiple posts associated with each user timeline, which is different from others collecting individual posts, however it doesn't conform to the definition of social media sessions despite being a closer approximation.

## 4.3. Cyberbullying annotation system

Despite slight variations in wording of the definition of cyberbullying, as well as different interpretations of the overlaps between cyberbullying and related terms (e.g. cyberaggression) there is an overall consistency in referring to the terms "repeatedly", "intended" and "power imbalance" when defining cyberbullying in the approaches where a rigorous definition is followed, as shown in Table 1.

Following the definition of what constitutes an act of cyberbullying, there have been predominantly two different means for labelling the data. Compared to labelling individual posts, the multimodality of social media sessions makes their labelling particularly challenging [39]. Before the annotation starts, it is important to carefully choose definitions of key terms that will inform annotators during the labelling process. For example, the concept of 'imbalanced power' might refer to one user being technically savvier than another, or a popular user abusing less popular users. Repeated cyberbullying can occur over time, such as retweeting/sharing profane comments multiple times [21]. When asking human annotators to determine whether a session constitutes a case of cyberbullying, it is important to incorporate information from all modalities, such as images and text-based comments [39].

*Crowdsourcing.* Most datasets have been annotated by crowd workers through online crowd-sourcing platforms [26, 27, 32, 29, 30]. This is indeed a challenging task for crowd workers who aren't necessarily trained on identifying cyberbullying, and therefore there is a risk that annotators may end up relying on other factors, such as the use of offensive words, beyond the fact of constituting an act of cyberbullying. To avoid this, [32, 33] followed an insightful approach of using previously annotated samples for selecting qualified annotators, i.e. those who got a minimum of labels right would qualify to conduct the rest of the annotation work.

*Research assistants.* In some cases, researchers have turned to trained annotators to conduct the manual annotation work, as is the case in [27, 28]. This has the advantage of having easier interaction with and control of annotators who are known to researchers, as well as the additional advantage of having been trained. It is however costly to recruit expert annotators as well as to train them for the annotation work.

## 4.4. SSCD datasets

Here we delve into more detail of the three datasets conforming to the two key criteria defined in the SSCD framework, i.e. collected based on social media sessions, and following a rigorous definition of cyberbullying that incorporates both its repetitive nature and the power imbalance, in addition to providing examples to annotators.

We describe these three datasets next, which were collected from Myspace, Instagram and Vine:

*Myspace [27].* This dataset consists of chat transcripts collected from MySpace.com. These chat transcripts are part of sessions whose structure is illustrated in Figure 5. These sessions take the shape of conversational threads, where the initial post introduces the discussion topic of the thread, following comments from others in the rest of the thread, which can often end up drifting from the original topic. Within these conversations, each post is considered a constituent part of the session, where a post can be lengthy, e.g. having multiple sentences or paragraphs. Because of the evolving nature of cyberbullying, the conversations were processed using a moving window of 10 posts to capture context. Each post consists of the user profile, date, and content.

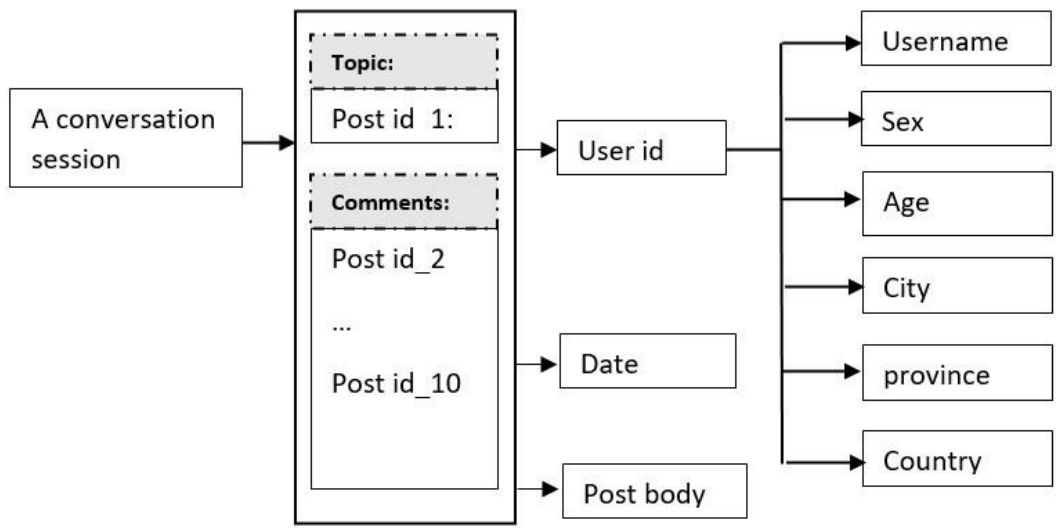

**Figure 5:** Illustration of the session structure in the Myspace dataset, where a session takes the form of a conversational thread including a post starting the conversation and followed by others commenting on it.

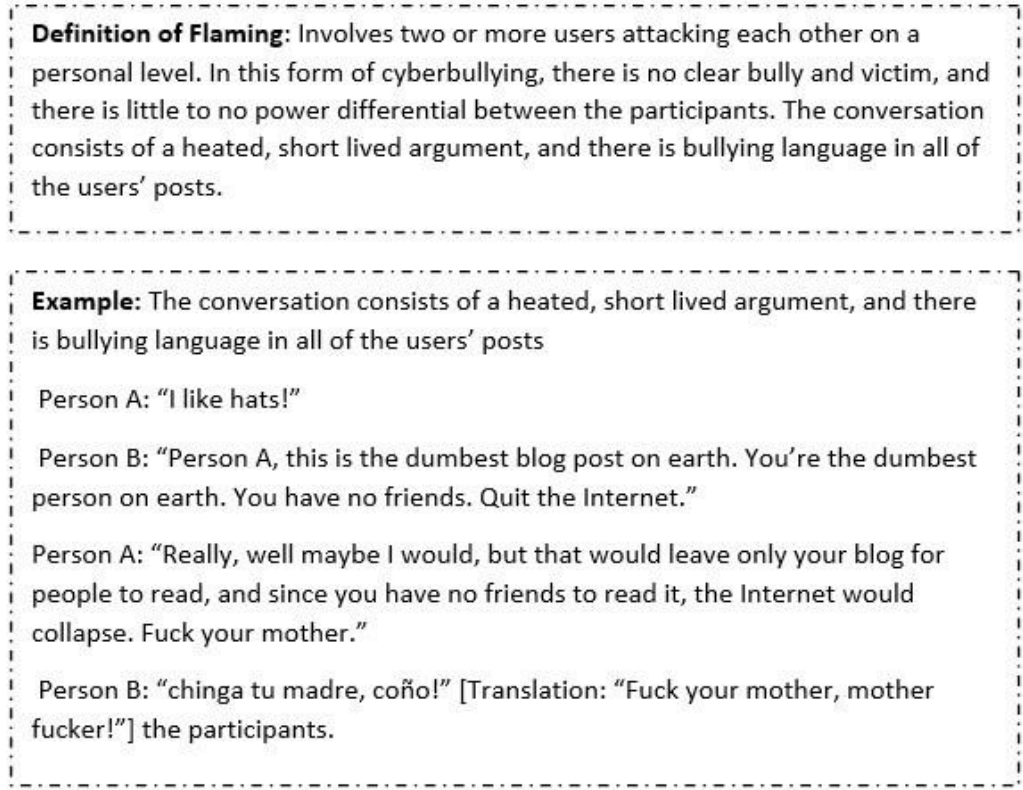

**Definition of Flaming**: Involves two or more users attacking each other on a personal level. In this form of cyberbullying, there is no clear bully and victim, and there is little to no power differential between the participants. The conversation consists of a heated, short lived argument, and there is bullying language in all of the users' posts.

**Example:** The conversation consists of a heated, short lived argument, and there is bullying language in all of the users' posts

Person A: "I like hats!"

Person B: "Person A, this is the dumbest blog post on earth. You're the dumbest person on earth. You have no friends. Quit the Internet."

Person A: "Really, well maybe I would, but that would leave only your blog for people to read, and since you have no friends to read it, the Internet would collapse. Fuck your mother."

Person B: "chinga tu madre, coño!" [Translation: "Fuck your mother, mother fucker!"] the participants.

**Figure 6:** A fragment of the MySpace annotation guidelines from [27].

Research assistants were provided with detailed annotation guidelines, of which a sample is shown in Figure 6. In these guidelines, authors divided cyberbullying into 9 categories, each of which provides detailed definitions and specific examples. Examples are not simply a single item of cyberbullying, but a cyberbullying session involving multiple user interactions. Annotators reviewed each window and were instructed to label whether or not it constituted a case of cyberbullying. Three annotators coded each item, after which the votes were aggregated through majority voting.

*Instagram [29].* Instagram is a social media platform where users can post images associated with comments, that others can like or reply to. In this dataset, authors collect each media object and its associated comments, which altogether make a social media session. Each media object contains the following information: media URL, media content, post time, caption and the number of likes/followed/shared. To facilitate the annotation work, authors filtered out sessions with fewer than 15 comments. Figure 7 shows a detailed structure of the sessions as stored in this dataset.

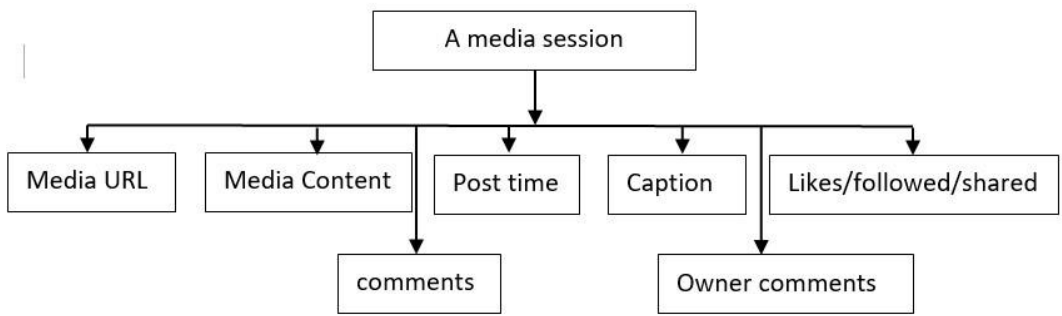

**Figure 7:** A media session structure of Instagram and Vine [29]

To ensure high quality annotation, authors restricted annotators to highly rated CrowdFlower workers. In addition, annotators were asked to answer a few test questions prior to taking on the annotation work, to further ensure that annotators were qualified. Annotators were given detailed guidelines including a definition of cyberbullying as well as a set of annotated examples. Annotators would also be disqualified if they completed the annotation work too quickly. In this case, each media session was annotated by five different workers, after which a final label was determined through aggregation. Figure 8 shows an example of the annotation interface used for this dataset.

*Vine[30].* Vine is a video-based online social network, where users can post videos and others can comment on them. This dataset was created by the same authors as the Instagram dataset and therefore followed a very similar approach to collect and annotate this dataset. The dataset in this case is made of media sessions with the same structure as the Instagram data, as shown in Figure 7, with the key difference being that they are initiated by videos rather than images. The annotation methodology is also identical to the Instagram dataset.

## 5. Approaches to Cyberbullying Detection

In this section, we will discuss existing cyberbullying detection techniques from the research papers under consideration, particularly delving into the approaches that consider social media sessions into their modelling. We discuss general trends in cyberbullying detection approaches first, moving on then to specific session-based approaches, with slight differences on how we discuss them:

*Trends in cyberbullying detection approaches.* Due to the large number of methods proposed for cyberbullying detection, these are generally tested on very different environments and settings, and hence it's difficult to compare them all together. Therefore, we focus on analysing the usage trends in types of algorithms, looking at how they are used in different scenarios.

*Session-based (SSCD) approaches.* Our study of SSCD modelling approaches is based on the unified framework of SSCD, we can more evenly analyse and compare the modelling motivations, techniques and corresponding performance, providing a comparison.

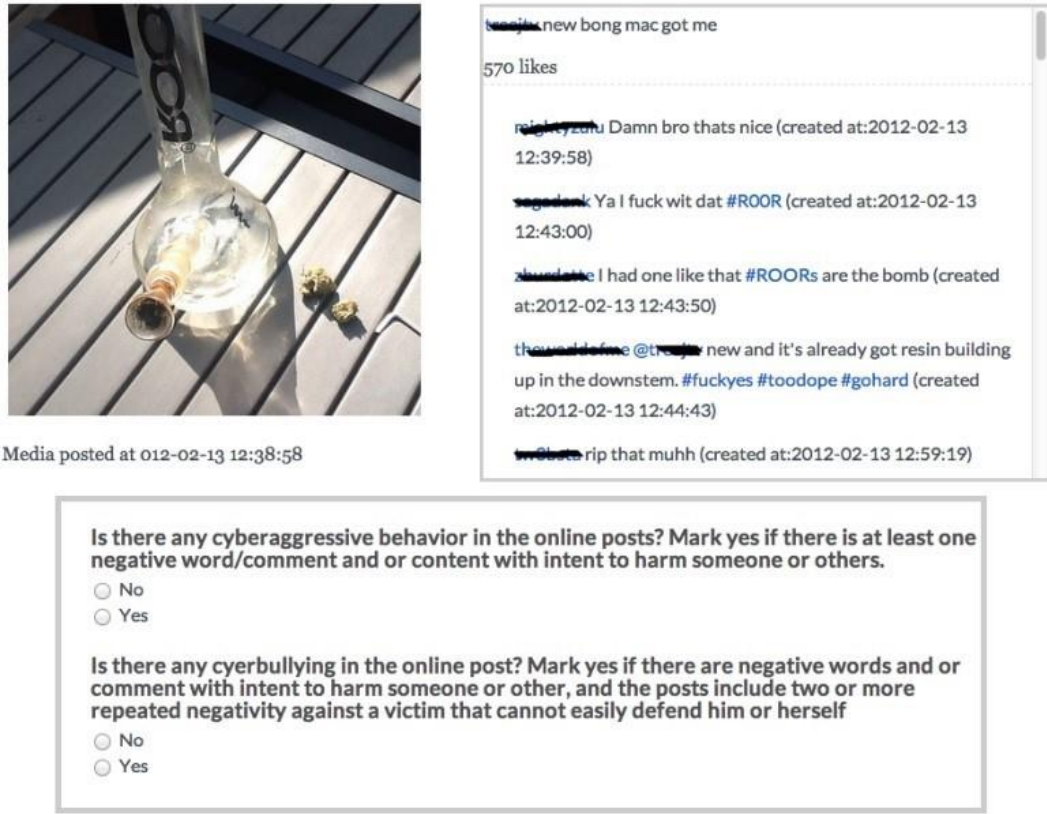

**Figure 8:** An example of the labeling study for Instagram datasets [29]

## 5.1. Trends in cyberbullying detection approaches

Table 2 shows a structured list of the different methods used in cyberbullying detection, showing also the list of research papers where each method has been employed. These methods can be grouped into three types, which we further discuss next: rule-based, machine learning and deep learning methods.

### 5.1.1. Rule-based methods

Rule-based approaches to cyberbullying detection have been studied for a long period, tracing back to 2008. Then, [40] used subjectivity analysis to design rules for extracting semantic information and keywords for cyberbullying detection. Another influential work using a rule-based approach was proposed by [42]. A framework-based method called lexical syntactic feature (LSF) is proposed to detect offensive content and predict whether a user is a bully, which is determined from a score generated by the model. A sentence-level offensiveness prediction is built, which uses lexical and syntactic features to calculate the offensiveness scores of content, and content-based features and writing style features are adopted to determine user offensiveness scores. Following a similar approach, [43] feed some person-specific references and multiple curse word dictionaries into a rule-based classifier to tackle the task. Their BullyTracer program used words from the selected dictionary, divided in three categories: insulting words, vulgar language and pronouns.

Rule-based approaches have proven to reach high accuracy in certain circumstances, but as posited by [42], they are severely limited to the predefined rules and fall short in their ability to generalise to new cases. Moreover, as they don't deal with social media sessions, no further context is used beyond single posts.

There has also been research [77, 78, 79, 61] analysing the distribution of "Bad words" in corpora, which is then used to identify the most prominent words that help generate a lexicon of "bad words" for cyberbullying detection.

| Approaches | Methods |
|---|---|
| Rule-based | 1. Semantic based [40, 41]<br>2. Lexical syntactic based [42]<br>3. Person-specific references and multiple cures word [43] |
| Machine learning methods | 4. Linear/Fuzzy Support Vector Machine [26, 44, 45, 46, 28, 47, 48, 49, 30, 50, 51, 33, 52, 53, 54]<br>5. K-nearest neighbours [26, 52, 54]<br>6. Logistic Regression [45, 30, 47, 55, 33]<br>7. Naive Bayes multinomial [45, 28, 30, 26, 33, 52, 56, 53, 54]<br>8. Conditional random fields [45]<br>9. Bayes Point Machine [57]<br>10. Stochastic Gradient Descent [52]<br>11. Random Forest [28, 30, 58, 26, 30, 59, 33, 52, 60, 56, 53]<br>12. Latent Dirichlet Allocation [30, 33]<br>13. Essential Dimensions of Latent Semantic Indexing [61]<br>14. AdaBoost [30, 30]<br>15. Reinforcement learning [62]<br>16. Time-Informed Gaussian Mixture Model [63]<br>17. Fuzzy logic and Genetic algorithm [64, 65, 52]<br>18. Participant-Vocabulary Consistency (PVC) using Alternating Least Squares [66] 16. Probabilistic Latent Semantic Analysis [48] |
| Deep learning | 19. Convolutional neural network [67, 58, 68, 69, 70]<br>20. (Bidirectional) Long Short-Term Memory [68, 69, 71]<br>21. Bidirectional Encoder Representations from Transformers [72]<br>22. Generative adversarial network [73]<br>23. Recurrent neural network [68]<br>24. Semantic-Enhanced Marginalized Denoising Auto-Encoder [74]<br>25. ConvNet [75]<br>26. Bi-GRU [76]<br>27. Hierarchical Attention Network [63] |

**Table 2**

Summary of cyberbullying detection methods and studies.

### 5.1.2. Machine learning methods

Machine learning models are, to date, the most widely used approaches. In addition, hybrid approaches as well as novel machine learning-based frameworks to solve specific complex problems are also being developed.

*Off-the-shelf machine learning algorithms.* In the early stages of using machine algorithms for cyberbullying detection, much of the research centred around the assessment of which machine learning algorithm performed best for the cyberbullying detection task [26, 45, 30, 80, 30] as well as around coming up with effective features to boost model performance through extensive feature engineering [57, 46, 47, 48, 55, 26, 59, 33, 52, 56, 81, 82, 83, 52, 54]. Classifiers such as Support Vector Machines (SVM), Naive Bayes (NB), Logistic Regression (LR), Random Forest (RF), Stochastic Gradient Descent, Bayesian Point Machines, Gradient Boosting, etc. have been extensively tested in various social media platforms such as Twitter, Facebook, MySpace, Formspring.me, Kongregate, Vine, etc. The first traceable work by using a machine learning method can be found in [84], where contextual features were also introduced for the first time for cyberbullying detection. This study builds on the hypothesis that the cyberbullying posts may be short and that detection can be supported through measuring their similarity with neighbouring posts. As well as in many other early cyberbullying detection studies, an SVM classifier is also used in this case.

*Hybrid approaches.* As machine learning models were found to be limited in dealing with some cases of cyberbullying, hybrid methods combining other strategies started to be proposed. For example, [44, 64, 28, 85] used fuzzy logic to determine importance scores of different classification models, depending on their advantages. [61] used Latent Semantic Indexing (LSI) [86], a commonly used method to match queries to documents, to match pre-defined cyberbullying query terms with relevant cyberbullying events by using second-order and higher-order word co-occurrences, which helped overcome synonymy and polysemy. [28] adopt an expert system (Multi-Criteria Evaluation Systems (MCES)) with other off-the-shelf machine learning algorithms [50] to extract graph-based features for each post, which is then fed to an SVM classifier. [87] used Latent Dirichlet Allocation (LDA) topic models to predict the probability that a given document belongs to a topic, subsequently using an NB classifier to assign posts to the different categories pertaining to different types of bullies.

| Type of features | Details | Number of papers |
|---|---|---|
| Content features | 1. Profanity | 23 |
| | 2. N-grams | 11 |
| | 3. Pronouns | 10 |
| | 4. Cyberbully keywords | 8 |
| | 5. TF-IDF | 7 |
| | 6. BOW | 3 |
| | 7. Skip grams | 1 |
| Sentiment Features | 8. Dictionary of words with sentiment | 13 |
| Social features | 9. Number comments | 3 |
| | 10. Number of Subscriptions | 2 |
| | 11. Number of uploads | 1 |
| | 12. Number of followers | 2 |
| | 13. Online time | 1 |
| | 14. Number of friends | 1 |
| | 15. Ego network | 1 |
| Media features | 16. ImageNet label | 1 |
| Writing style features | 17. Length of messages | 3 |
| | 18. Count/ratio of emoticons | 2 |
| | 19. Spelling | 2 |
| | 20. Capitalisation | 2 |
| | 21. Parts-of-speech tagging | 1 |
| | 22. The length of the text | 1 |
| | 23. Number of pronouns | 1 |
| Demographic features | 24. Age | 5 |
| | 25. Gender | 3 |
| | 26. Location | 1 |

**Table 3**
List and paper count for different features used in cyberbullying detection.

[49] built a method inspired by the Multiple Sequence Alignment (MSA) method, which is commonly used in computational biology for identifying conserved regions of similarity among raw molecular data. They converted cyberbullying data into string sequences for revealing conserved temporal patterns or slight variations in the attacking strategies of bullies.

*Novel frameworks.* Given the complexity of cyberbullying datasets, researchers started to develop multi-layer components or multi-model combinations. [51] created a bullying severity identifier composed of multiple fuzzy logic systems.

Aiming to reduce the number of features used to classify comments and the scalability of online detection, [20] divided the binary classification task into two tasks under a novel framework. One aims to determine if there is an incident of cyberaggression in the comment stream, and the other aims to introduce "repetitiveness" as a threshold to detect session-level cyberbullying.

*Feature engineering.* Almost all of the supervised learning approaches go through careful feature engineering. In Table 3, we list the set of features that have been used across different studies, along with their associated count. We can see that the most popular features are the *content-based* ones, such as cyberbullying keywords from lexica, topic-based profanity, pronouns, n-grams, Bags-of-words (BOW), Term Frequency Inverse Document Frequency (TF-IDF) etc. Especially, the profanity lexicon is widely used as a cue to detect potential cyberbullying events. However, researchers have pointed out that solely using content-based features can be limited in capturing other inherent characteristics of cyberbullying such as personalisation, contextualisation and diversity, which motivated the use of other features [77].

Sentiment, social and writing style features are also widely used, whereas media-based and demographic features are rarer:

*Sentiment features.* Most researchers generally use the phrase, keywords and symbols as an indicator of the sentimental expression in a post [88, 89, 90, 91, 92, 21]. While sentiment features are popular in cyberbullying, they also tend to be insufficient to be used alone and are generally used jointly with other features.

*Demographic features.* Including the use of gender-specific, age-specific or location vocabularies. [3] also noted that features inferred from author profiles can be effectively used to improve performance.

*Social features.* Which include features such as followers or online time, tend to be specific to each social media platform and hence more difficult to generalise across platforms, however have also proven to be effective in boosting the performance in specific environments [4, 49].

*Writing-style features.* Including features such as "pronoun + profanity" [88], document length [4], word capitalisation [44] and spelling [3], which have shown to be good predictors of a user's likelihood of engaging in abusive behaviour in social media [84].

*Media features.* As a rather unique and seldom used feature, image-related features were used in the study by [71].

### 5.1.3. Deep learning methods

In recent years, there has been a clear shift from the use of machine learning models to an increasing use of deep learning models. With the use of deep learning architectures, other more sophisticated features such as polymorphism, dynamism, hierarchical, and interactivity have also been studied.

Deep learning models have been used to improve representations that are then fed to machine learning algorithms or used in shallow neural networks [93, 60, 94, 95]. For example, [60] trained a word embedding model that is based on the word2vec skip-gram model for exploring better sentence embeddings, with an RF used for the final classification. Semantic-Enhanced Marginalized Denoising Auto-Encoder [93] (smSDA) was developed via semantic extension of the stacked denoising autoencoder. The semantic extension consists of semantic dropout noise and sparsity constraints, where the semantic dropout noise is designed based on domain knowledge and the word embedding technique. Linear SVM is then applied to the new feature space.

Deep learning models have also been stacked into hierarchical structures that mirror the complex data structure. [58, 68, 96, 97, 69, 62, 75, 71, 70]. Convolutional Neural Network (CNN), Bidirectional Long Short-Term Memory (BLSTM), Gated Recurrent Unit (GRU), Long Short-Term Memory (LSTM), and Recurrent Neural Networks (RNN) are the most commonly used deep network architectures for these purposes. [58] were the first to use a CNN to transfer from an image classifier to a cyberbullying classifier. Among these deep learning studies, [98] proposed using semantic domain knowledge (demographics, text and social features) to drop out noise and to increase hidden features in the word embedding using stacked deep learning techniques. They then used them in a classification layer for making the whole model more discriminative. [68] analyse cyberbullying detection on various topics across multiple social media platforms using a deep learning model with transfer learning. [99] proposed a double-balanced framework to tackle two important issues: variant contribution and imbalance datasets. [10] used three different types of word embeddings (Word2vec, GloVe, ELMo) that were tested as inputs and coupled with different deep learning architectures.

## 5.2. SSCD modelling methods

In this section, we focus on modelling approaches for cyberbullying detection based on the SSCD framework. Existing efforts extend text-based analysis to session-based analysis, an extension that is based on the inherent hierarchy of conversations (e.g. word forming comments, dialogue comments), multimodal data (e.g. text, location, images, etc.) and user interactions.

*Inherent hierarchies with attention.* Emerging literature identifies cyberbullying as repetitive temporal acts rather than one-off incidents. Thus, modelling the hierarchical structure of social media sessions and the temporal dynamics of cyberbullying in online social network sessions are key distinctive characteristics of this approach, which are generally considered through three different yet complementary means:

- The hierarchical network structure is adapted to reflect the structure of a social media session.

- Instead of relying on handcrafted features, they leverage an attention mechanism to automatically capture word level and sentence-level hidden embeddings. They then weight them to form a more representative document level representation.

- The interval of time between two adjacent comments is considered in a hierarchical network.

HANCD [69] and HENIN [100] can be viewed as two presentive instantiations of this approach. HANCD consists of two levels of Hierarchical Attention Network (HAN): one at the word level and the other at the comment level. These two HANs can capture the differential importance of words and comments in different contexts. Then the bidirectional GRU is employed to capture the sequence of contents. HENIN focuses more on learning various interactions between heterogeneous objects displayed in social media sessions. A comment encoder is created to learn the representations of user comments through a hierarchical self-attention neural network so that the semantic and syntactic cues of cyberbullying can be captured. A post-comment co-attention mechanism learns the interactions between a posted text and its comments. Moreover, two graph convolutional networks are leveraged to learn the latent representations depicting how users interact with each other in sessions, and how posts resemble each other in terms of content.

*Multimodal models.* Social media sessions are often multimodal (e.g., image, video, comments, time). Hence, there has also been research in making the most of this diversity of modalities. [101] used encoder denoising techniques and constraints on sparse hidden features. Regarding the method of integrating presentations, a straightforward approach to encode multi-modal context is to simply concatenate the raw feature vectors of each modality (e.g., locations, comments, images, timestamps) [83]. However, this method overlooks both structural dependencies among different social media sessions and cross-modal correlations among different modalities. MMCD [83] proposed to train a model based on three different components: (i) Topic-oriented bidirectional long-short term memory (BiLSTM) model with self-attention, (ii) comment-based Hierarchical Attention Network(HAN) to focus on word-level and comments-level characteristics, and (iii) visual embeddings to encode different types of modes. Then they integrated them into a hierarchical attention network to capture hierarchical relationships. XBully [102] is another presentive model, which reformulates multimodal social media data as a heterogeneous network and then aims to learn node embedding representations upon it. In contrast to simply concatenating the raw multi-modal feature vectors of each modality, multiple learned nodes embedded into the resultant heterogeneous network may generate more complex and specific presentations.

*User interaction extractors.* Cyberbullying often takes place throughout a series of interactions on social media platforms. Therefore, approaches incorporating sequences of user interactions have also been studied. For example, [41] use a rule-based classifier to tag a conversation session into a sequence of sentiment words to reflect user interactions. [20] proposed a novel algorithm called CONcISE, which reduces the number of classification features used for detecting cyberbullying. The main idea is to feed the sequential aggression detection results of each session into the next high-level cyberbullying detection classifier. Through the so-called Time-Informed Gaussian Mixture Model (UCD), [99] proposed an Unsupervised Cyberbullying Detection method, which incorporates comment inter-arrival times for social media sessions, allowing the use of the full comment history to classify instances of cyberbullying. [103] used a graph neural network for modelling topic coherence and temporal user interactions to capture the repetitive characteristics of bullying behaviour, thus leading to better predicting performance.

*Performance comparison.* Most of the modeling methods mentioned above were experimented on two SSCD datasets: Instagram and Vine. The consistency of how these models have been evaluated facilitates comparison between model performances, which we show in Table 4. Still, it is worth noting that, in addition to the differences in the proposed models, there may be other differences in the preprocessing of the data.

In the comparison of the six models corresponding to the three modelling methods, none of the models achieves consistently the best performance across the two datasets. XBully achieves the best performance among the six models on the Instagram dataset, and MMCD outperforms four other models on the Vine dataset. These two models both adopted multimodal modelling strategies, which suggests that they are promising methods for SSCD. Another interesting observation we make is that the overall performances on Vine are lower than on Instagram, even if the session structure of both datasets is the same.

| Approach | Model | Instagram | Vine |
|---|---|---|---|
| Inherent hierarchies with attention | HANCD [69] | 0.851 | N/A |
| | HENIN [100] | 0.838 | 0.676 |
| Multimodal model: | LSTM + context2vec features [71] | 0.85 | N/A |
| | MMCD [101] | 0.86 | 0.841 |
| | XBully [102] | 0.878 | 0.804 |
| User Interaction Extractors: | TGBully [103] | 0.81±0.02 | 0.69±0.02 |

**Table 4**

Performance comparison of SSCD models on two SSCD datasets: Instagram and Vine.

| Model | Instagram | Vine | Average |
|---|---|---|---|
| MMCD | 0.86 | 0.84 | 0.85 |
| XBully | 0.88 | 0.80 | 0.84 |
| BERT [104] | 0.77 | 0.83 | 0.80 |
| ROBERTA [105] | 0.85 | **0.89** | **0.87** |
| MPNET [106] | 0.85 | **0.87** | **0.86** |
| LONGFORMER [107] | 0.77 | **0.86** | 0.82 |
| T5 [110] | 0.79 | **0.94** | **0.87** |
| XLNET [108] | 0.83 | **0.87** | 0.85 |
| ELECTRA[112] | 0.83 | **0.88** | **0.86** |
| DISTILBERT [109] | 0.82 | **0.87** | 0.85 |
| BERTWEET [111] | 0.76 | 0.43 | 0.60 |

**Table 5**

Performance of pre-trained language models and state-of-the-art SSCD models.

# 6. Benchmark Experiments with Pre-trained Language Models

In this section, we focus on investigating and benchmarking the effectiveness of a range of models. While not all the SSCD models presented in Section 5.2 are available for reproducibility, we present results for two of them: MMCD and XBully. In addition, we test a range of large pre-trained language models: BERT [104], ROBERTA [105], MPNET [106], LONGFORMER [107], XLNet [108], DISTILBERT [109], T5 [110] and BERTWEET [111]. We test all these models on two datasets: Instagram and Vine.

To set up these experiments, we follow the same preprocessing method as [103]. For the implementation of pretrained language models, we use HuggingFace. The number of training epochs used is 5. We split the data in stratified samples of 80% and 20% for training and testing.

Table 5 shows the Macro-F1 scores of all the models tested. We observe that both MMCD and XBully are competitive models outperforming all pre-trained language models on the Instagram dataset. However, on the Vine dataset, the majority of the pre-trained models, except BERT and BERTWEET, outperform both MMCD and XBully. If we look at the average performances across both datasets, four pre-trained models, namely ROBERTA, MPNETM, T5 and ELECTRA, show better generalisability than MMCD and XBully.

These experimental results demonstrate that pre-trained language models can be strong, competitive models for Social Media Session-Based Cyberbullying Detection. Still, the differences in performance we observe across both these datasets call for the implementation of more generalisable models that can perform well across different platforms and datasets. This in turn requires creation and release of additional datasets, ideally from different social media platforms, to further study the generalisability of models beyond these two platforms.

# 7. Best Practices for Creating SSCD-based Datasets

Existing resources, including both models and datasets, are useful to learn about and design best practices for SSCD dataset creation. Informed by this prior research, we provide recommendations in the three key steps for dataset creation: (i) social media platform selection, (ii) session-based data collection, and (iii) cyberbullying annotation.

## 7.1. Social media platform selection

The selection of a suitable social media platform for the creation of a dataset should be motivated by the problem and research questions at hand. Where applicable, this motivation can be further strengthened through interdisciplinary collaborations that can help shape stronger and more comprehensive research questions. Still, one of the aspects to take into account is the inherent diversity of cyberbullying and the diverse set of ways in which it is manifested, which also complicates collection of a dataset encompassing this diversity. Hence, it's also important to clearly define what kinds of cyberbullying events a platform is expected to deliver.

## 7.2. Session based Data collection

The social nature of cyberbullying requires collection beyond simple textual posts, including also its hierarchical structure (i.e. words form comments, comments from conversations), multimodal data (i.e. text, location, user profile, etc.) and evolving user interactions. Hence, datasets should also incorporate this hierarchy, multimodality and user interactions.

This in turn enables more in-depth and careful implementation of models leveraging the social nature of cyberbullying. For example, session-based investigations can provide valuable insights into the power imbalance between the bully and the victim, which is only likely to manifest across the entire session, and may not be observed when looking only at individual texts. The repetitive nature of cyberbullying can be captured by the sequence of comments in a conversation. Examining the hierarchy of social media sessions also enables the model to distinguish the importance of media objects in the session. Thus, session-based detection of cyberbullying opens up promising research directions for identifying, understanding, and ultimately preventing cyberbullying in the real world.

Ensuring that enough cyberbullying cases are captured in a dataset is another challenge, because of the 'rarity' of cyberbullying events if we look at all the content in a social media platform. Many sampling strategies, such as those based on keywords, can introduce a bias in the data selection, and therefore designing a careful data sampling strategy is crucial. At least two promising research directions can help mitigate this bias: (i) intentionally incorporating synthetic yet realistic "perturbations", with the aim of diversifying the content while also preserving its real-world nature, and (ii) careful collection of negative samples, once the positive samples are collected through a carefully designed strategy.

### 7.3. Cyberbullying annotation

Data annotation is a time-consuming and labour-intensive process. So far, crowdsourcing platforms have been the prevalent option for researchers to annotate datasets. Crowdsourcing has multiple advantages, such as ensuring the diversity and scope of the overall workforce, however it comes at the cost of having a likely untrained set of annotators for what can be considered a relatively challenging annotation work. If crowdsourcing platforms are used, it is advisable to carefully design the annotation guidelines, with sufficient examples including 'edge cases', and to ensure that annotators are qualified, for example through test questions prior to starting the annotation. Previous literature has highlighted the difficulty of distinguishing various types of misconduct [38].

Data labelling is particularly challenging due to the multimodality of social media conversations. When asking annotators to determine whether a conversation is cyberbullying, it is important to integrate all available information with different forms of data components, such as images and text-based comments.

Still, annotation through trained annotators is ideal for a challenging task like cyberbullying, with the main challenges of having access to a set of qualified people, as well as its associated cost.

## 8. Open Challenges

Social media session-based cyberbullying detection presents multiple challenges and promising opportunities that differ from single-text based cyberbullying detection tasks. In this section, we will highlight two open challenges that emerge from our investigation of the subject, related to datasets and models:

*Improving the quality of datasets and the clarity in reporting about them.* It is often the case that not enough information is reported on how datasets have been created, and how the different underlying factors (i.e. repetition and power imbalance) have been considered, if they have. Dictionaries of "bad words" are often used for the data collection, which enables collection of certain types of cyberbullying but misses other cases where those keywords aren't present. This in turn limits the generalisability of the models tested on those datasets, and therefore studying improved data collection strategies should be a priority. In creating cyberbullying datasets, researchers should also avoid conflation with the related concepts of toxicity and hate speech, which differ for example in the fact that they are not necessarily repetitive.

*Increasing the reliability and reproducibility of models.* Not all cyberbullying models are reported with sufficient details; where the code of these models isn't published, it also means that they are not replicable because the level of detail is insufficient. In order to further research in cyberbullying detection, it is crucial to enable reproducibility of existing models, so that researchers can build upon and improve existing models. Likewise, it is also important that research in cyberbullying detection considers more than a single dataset in their studies, which enables evaluating the generalisability of models demonstrating competitive performance not only on a single dataset.

## 9. Conclusion

In this survey paper, we review existing approaches to cyberbullying detection, with a particular focus on session-based cyberbullying detection, for which we define the Social media Session-based Cyberbullying Detection framework (SSCD) made of four key components. By going through the research challenges and progress on the four components of the SSCD framework, we review existing research in cyberbullying detection through model and dataset creation, particularly delving into those dealing with social media sessions.

Through our review, we also highlight the importance of considering two of the inherent characteristics of cyberbullying when designing dataset creation and experiments, i.e. repetition and power imbalance. Our review discusses existing dataset and models, presents a set of comparative, benchmark experiments evaluating state-of-the art models on SSCD datasets, as well as posits a set of suggestions for future research when it comes to dataset and model creation.

Where SSCD is an emerging research trend, our survey provides a valuable reference for those studying the problem.

Peiling Yi, A research student in the Cognitive Science Research Group, Queen Mary University of London. A large part of her research interest falls in the intersection of Transfer learning and Text classification. Currently, the main research interest is in cyberbullying detection across different social media platforms.

Dr. Arkaitz Zubiaga, Lecturer at the Queen Mary University of London, where he leads the Social Data Science lab. His research interests revolve around linking online data with events in the real world, among others for tackling problematic issues on the Web and social media that can have a damaging effect on individuals or society at large, such as hate speech, misinformation, inequality, biases and other forms of online harm.